\documentclass[11pt,a4paper]{article}
\usepackage{booktabs} % For formal tables
\usepackage{bm} %bold equations
\usepackage{subcaption} % For subfigures and subcaptions
\usepackage{multirow}
\usepackage{pgf}
\usepackage{hyperref}
\usepackage{natbib}
\usepackage{amsmath}

\title{Mamba base PKD for efficient knowledge compression}
\author{
José Medina\textsuperscript{1} \and 
Amnir Hadachi\textsuperscript{1} \and 
Paul Honeine\textsuperscript{2} \and 
Abdelaziz Bensrhair\textsuperscript{3}
}

\begin{document}
\maketitle
% Then after \maketitle, add:
\begin{center}
\begin{minipage}{0.8\textwidth}
\small
\textsuperscript{1}ITS Lab, Institute of Computer Science, University of Tartu, Estonia \\
\textsuperscript{2}Litis, Université de Rouen, France \\
\textsuperscript{3}Litis, INSA de Rouen, France \\
\texttt{\{joseluis,hadachi\}@ut.ee}, \texttt{paul.honeine@univ-rouen.fr}, \texttt{abdelaziz.bensrhair@insa-rouen.fr}
\end{minipage}
\end{center}

\begin{abstract}
Deep neural networks (DNNs) have remarkably succeeded in various image processing tasks. However, their large size and computational complexity present significant challenges for deploying them in resource-constrained environments. This paper presents an innovative approach for integrating Mamba Architecture within a Progressive Knowledge Distillation (PKD) process to address the challenge of reducing model complexity while maintaining accuracy in image classification tasks. The proposed framework distills a large teacher model into progressively smaller student models, designed using Mamba blocks. Each student model is trained using Selective-State-Space Models (S-SSM) within the Mamba blocks, focusing on important input aspects while reducing computational complexity. The work's preliminary experiments use MNIST and CIFAR-10 as datasets to demonstrate the effectiveness of this approach. For MNIST, the teacher model achieves 98\% accuracy. A set of seven student models as a group retained 63\% of the teacher’s FLOPs, approximating the teacher's performance with 98\% accuracy. The weak student used only 1\% of the teacher’s FLOPs and maintained 72\%  accuracy. Similarly, for CIFAR-10, the students achieved 1\% less accuracy compared to the teacher, with the small student retaining  5\% of the teacher's FLOPs to achieve 50\% accuracy. These results confirm the flexibility and scalability of Mamba Architecture, which can be integrated into PKD, succeeding in the process of finding students as weak learners. The framework provides a solution for deploying complex neural networks in real-time applications with a reduction in computational cost.
\end{abstract}

\section{Introduction}
A preliminary version of this work was presented as a short poster titled 
\textit{``Mamba-PKD: A Framework for Efficient and Scalable Model Compression in Image Classification''} at The 40th ACM/SIGAPP Symposium on Applied Computing (SAC ’25) \footnote{A preliminary version of this work appeared as a conference poster: 
 \href{https://doi.org/10.1145/3672608.37078877}{https://doi.org/10.1145/3672608.3707887}}, This version extends the previous work by providing a detailed description of the methodology, additional experiments, and a more comprehensive discussion of results.\\

\begin{figure}
\centering
\includegraphics[width=\linewidth]{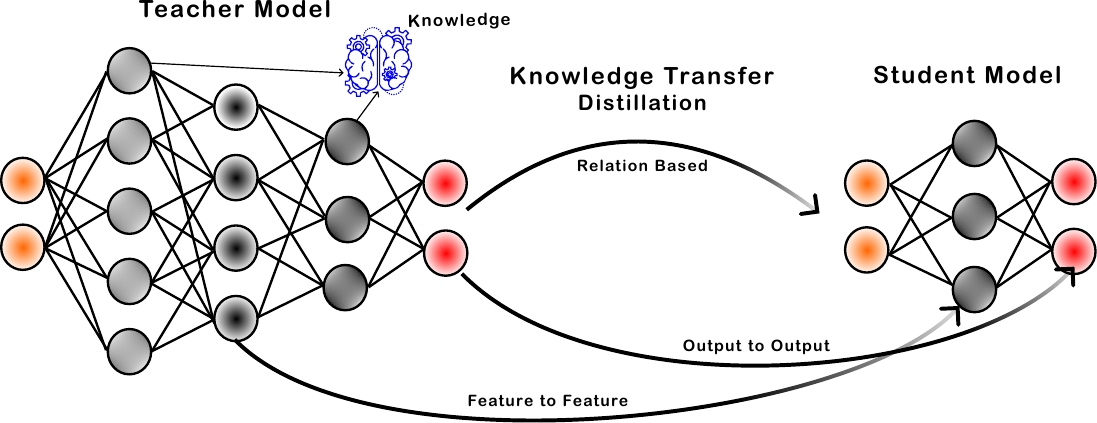}
\caption{Teacher-student framework for knowledge distillation, where knowledge can be transferred based on teacher-student outputs, features, and pure relations.}
\label{fig:teacher-student}
\end{figure}

Deep Neural Networks (DNNs) are large-scale models extensively used in natural language processing, image processing, and speech recognition tasks. These models are typically over parameterized to ensure generalization and effective feature extraction~\citep{gou2021knowledge}, but their size demands significant computational resources and generates challenges for real-time applications. Researchers have explored methods to address these issues by transferring the knowledge encapsulated in unwieldy models to more lightweight neural networks. The approach is known as Knowledge Distillation (KD) and involves compressing DNN architectures by transferring knowledge from a complex teacher model to a simpler student model without compromising accuracy and reliability~\citep{hinton2015distilling}. The student model learns to mimic the teacher's behavior by focusing on its outputs and neuron connections rather than raw data, reducing computational requirements while maintaining competitive performance~\citep{wang2021knowledge}. KD encompasses the design of three components: knowledge modeling, distillation algorithms, and teacher-student architectures \citep{gou2021knowledge}, as illustrated in Figure~\ref{fig:teacher-student}.\\

Extending the principles of KD, recent research has introduced Progressive Knowledge Distillation (PKD), which transforms the knowledge transfer process from a rigid student architecture to a progressive and dynamic one~\citep{dennis2023progressive}. Instead of distilling knowledge all at once, PKD allows the student model to grow incrementally and refine its predictions over time by breaking down the teacher model into a series of smaller student models, or weak learners, that collaboratively approximate the teacher's behavior. This progressive approach offers a flexible trade-off between inference cost and model accuracy, making it particularly beneficial for scenarios that require adaptive inference. Moreover, PKD can assemble the predictions of each student in parallel, enhancing coarse predictions by summing the contributions of different students without the need to reevaluate the entire model. This method not only improves efficiency but also adapts to varying resource constraints and performance requirements.\\

Related to enhancing efficiency, new approaches based on State Space Models (SSMs) make significant contributions. These mathematical frameworks are ideal for modeling systems whose state evolves over time based on its previous states and current inputs, by considering simple matrix multiplications~\cite {gu2021efficiently}. SSMs serve as a foundation for capturing temporal dependencies, in tasks such as pixel sequences and time series in computer vision~\citep{nguyen2022s4nd,lei2024dvmsr}. However, traditional SSMs represent the entire system, treating all incoming information equally important, making them effective for modeling data but impractical for large sequences. \citet{gu2023mamba} handled this inefficiency by introducing innovations in an architecture called Mamba. The central concept of this architecture is a Selective-State Space Model (S-SSM), which process information based on the current input, focusing on relevant information while discarding irrelevant data. This selective process reduces unnecessary computation, ensuring that resources are allocated to meaningful tasks.\\

The present work introduces a novel approach to model compression by integrating Mamba architecture within a PKD framework. This progressive distillation process enables the model to scale flexibly according to available resources and specific application needs. Mamba architecture significantly enhances efficiency by leveraging S-SSM, which focuses only on relevant inputs while discarding irrelevant data. This selective processing allows Mamba to act as an efficient backbone for progressively distilled models; for instance, in our results for MNIST, a group of seven student models trained progressively retained only 63\% of the teacher’s FLOPs while maintaining the same teacher accuracy of 98\%. An important feature of this framework is also the hardware-aware Mamba algorithm, which can manage several student models in parallel as they progressively refine their coarse predictions. By distributing computations across students, the overall inference time is reduced to that of the most complex student, plus a minimal aggregation step, which is much faster than the inference time required by the teacher model alone. In CIFAR-10, for example, our framework achieves 50\% accuracy from teachers using only 5\% of the teachers' FLOPs in one student. This makes the framework particularly suitable for real-time applications and resource-constrained environments, where balancing efficiency and accuracy is crucial.\\

The primary contributions of this work are: \begin{itemize}
    \item \textbf{Proposed Training Pipeline:} A pipeline combining PKD with Mamba architecture to progressively construct and train student models from a teacher, ensuring incremental refinement.
    \item \textbf{Hardware-Aware Parallel Processing Algorithm:} Efficient parallel management of student models, reducing inference time for complex data sequences and multiple students.
    \item \textbf{Flexible and Efficient Mamba Blocks:} Student models are trained using Mamba blocks, which selectively handle important features while minimizing model size. This framework offers flexibility, facilitating the student's training by adjusting levels of complexity and computational demands.
    \item \textbf{Validation on Benchmark Datasets:} Experiments on MNIST and CIFAR-10 demonstrate the framework’s scalability and reduced computational cost.
\end{itemize}

This paper is structured as follows: Section~2 provides an overview of works related to KD, PKD, and SSM. Section~3 delves into the theoretical foundation for combining Mamba architecture within PKD. %and highlighting their role in enhancing model efficiency. 
Section~4 describes the experimental setup, including the datasets and hyperparameters. Section~5 presents the results of combining Mamba with PKD. Finally, Section~6 discusses the implications of the findings and outlines potential future work.

\section{Related Work}
Developing efficient deep learning models has driven interest in techniques like Knowledge Distillation, Progressive Knowledge Distillation, and efficient architectures like Mamba. Below, we explore key areas where these techniques have evolved and discuss recent research trends.
\subsection{Knowledge Distillation}
Knowledge distillation has been extensively studied as a model compression technique that allows a smaller student model to learn from a larger teacher model~\cite{gou2021knowledge,han2015deep} (see Figure~\ref{fig:teacher-student}). Traditional KD includes transferring the knowledge to a student model from a teacher model's output~\citep{chen2017learning}, intermediate layers \citep{romero2014fitnets}, or only the relationships between different layers and data samples~\citep{yim2017gift,passban2021alp}. These techniques lead to computational savings while preserving high accuracy. However, one limitation of traditional KD is the performance drop when there is a large capacity gap between the teacher and student models because the student cannot effectively represent key features of the teacher. This limitation has been addressed by various methods \citep{romero2014fitnets, yim2017gift,zhang2019your, passban2021alp} that modify distillation algorithms or employ multi-stage learning.\\

To overcome these challenges, hierarchical or layer-wise distillation approaches were proposed, where knowledge is progressively transferred across multiple layers of the teacher to the student. \mbox{FitNets},  introduced in \citep{romero2014fitnets}, transfer intermediate representations from the teacher model to the student, improving the student’s ability to learn fine-grained features progressively. Similarly, self-distillation \citep{zhang2019your} trains a model by distilling knowledge into itself at different stages, a concept related to progressive distillation where multiple students incrementally improve performance \citep{dennis2023progressive}.
\subsection{Progressive Knowledge Distillation}
\begin{figure}
\centering
\includegraphics[width=\linewidth]{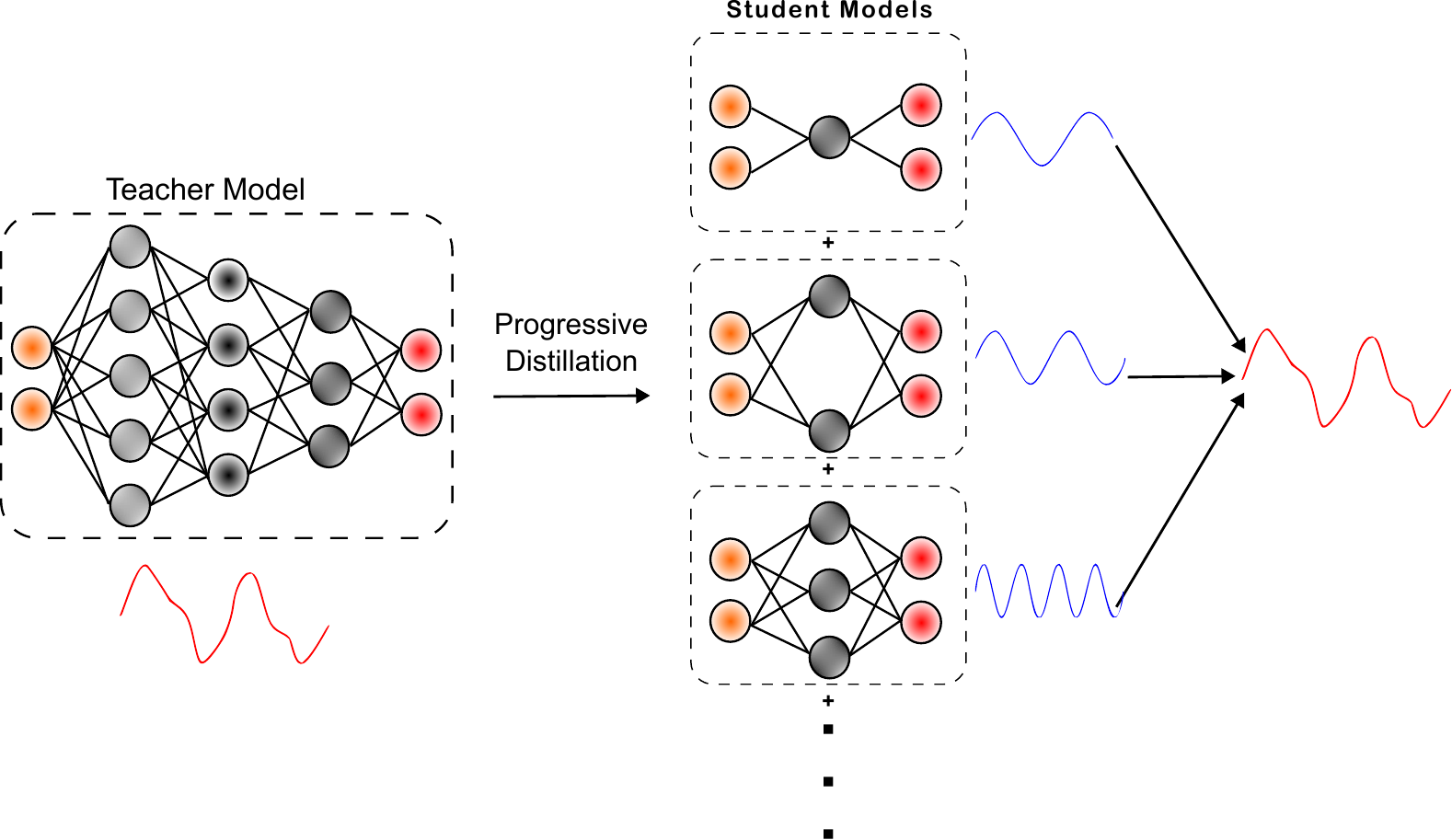}
\caption{Progressive Knowledge Distillation, the knowledge from a complex architecture teacher is split among different simple architecture students, who are trained to progressively improve a coarse prediction.}
\label{fig:progressive}
\end{figure}

Building on traditional distillation methods, PKD incrementally transfers knowledge to improve the efficiency and scalability of the distillation process. In \cite{dennis2023progressive}, the authors proposed the B-DISTIL algorithm, which decomposes a large teacher model into an ensemble of smaller student models, each capable of refining predictions as more models are evaluated (see Figure~\ref{fig:progressive}). This way of distillation enables a flexible trade-off between inference time and accuracy, making it particularly effective for on-device inference, where computational resources are limited.\\

This method also allows for early-exit and anytime inference, where the student models can deliver predictions after evaluating partial subsets of the model, reducing the need for full evaluations in real-time settings. Techniques like these are closely related to dynamic and adaptive neural networks~\citep{song2022spot}, where the complexity of the model can adjust dynamically based on resource availability.
\subsection{State Space Models and Selectivity}
SSMs have long been used to capture temporal dependencies in sequence data, but they often face inefficiencies, particularly when applied to large or complex sequences, as their uniform processing of all incoming information can lead to performance bottlenecks \citep{gu2021efficiently}. To address complex sequences, the Mamba Architecture described in \citep{gu2023mamba} was introduced with Selective-State-Space Models, which selectively process only the most relevant information from the input data while discarding irrelevant information. This innovation enables Mamba to drastically reduce computational requirements by allocating resources more efficiently, focusing computation only on critical elements. \\

Unlike Transformer-based models \citep{vaswani2017attention}, which rely on attention mechanisms and MLP blocks, Mamba's architecture uses a single, unified SSM block, offering a more computationally efficient alternative for sequence modeling.\\%This makes Mamba particularly useful for real-time applications that require both scalability and speed.
\subsection{Dynamic Neural Networks and Hardware-aware Parallelism}
The concept of dynamic and adaptive neural networks has also gained traction \citep{zhang2019your,song2022spot}, where models adjust their complexity and depth during inference based on resource availability. Adaptive neural networks have been proposed to enable models to exit early when a high-confidence prediction is reached, reducing unnecessary computation when a full evaluation is not needed \citep{bolukbasi2017adaptive}. This approach complements PKD \citep{dennis2023progressive}, where student models progressively refine their performance, allowing the evaluation to stop early once the prediction is sufficiently accurate.\\

Moreover, Mamba's hardware-aware parallelism \citep{smith2022simplified} is specifically designed to enhance efficiency on edge devices and resource-constrained platforms. By leveraging selectivity in SSM, the use of convolution is constrained, so scans (hardware-aware) optimize the inference time to be linear to the length of input sequences. This hardware-aware algorithm also uses GPU memory hierarchy to control a fast data transfer \citep{gu2023mamba}, which can be used to transfer data and coordinate inference enhancement when working with several students running in parallel, as in PKD. 

\section{Theory and Methodology}
In this section, we explain the theoretical foundations of integrating PKD with the Mamba Architecture. Based on these concepts, we outline the methodology of the proposed architecture.
\subsection{Knowledge Distillation Foundation}
The simple way to explain KD is to follow a process governed by a distillation loss, which balances two components: hard and soft labels. For hard labels, the loss function $L_{hard}$ is conventionally computed as a cross-entropy loss between the student predictions and the true labels. For soft labels, a loss function ${L_{soft}}$ is computed between student outputs and teacher soft outputs, applying temperature before the softmax function. The distillation loss $L_{KD}$ can be expressed as:
\begin{equation}
    L_{KD}=\alpha L_{hard}+(1-\alpha)L_{soft}
    \label{eq:KDLoss}
\end{equation}
where \( \alpha \) is a weighting factor that balances the two components.\\

In PKD, the teacher model’s knowledge is distilled into a sequence of smaller student models, also referred to as weak learners. The students are trained in the following way: 1) The first student model learns a coarse representation of the teacher's knowledge; 2) Successive students refine their predictions based on intermediate layers of the teacher and earlier students; 3) The final model aggregates predictions from all students or uses the last, most accurate student for full inference. This process is presented as the B-DISTIL algorithm \citep{dennis2023progressive}, which frames distillation as a two-player zero-sum game. In this game, the teacher model acts as one player (the maximizer), producing distributions over the training data. In contrast, the student models act as the second player (the minimizers), iteratively learning from the teacher’s predictions. The primary goal is to find student models that can effectively approximate the teacher model’s predictions while progressively improving upon earlier student models. \\

At a round \( t \) (looking for weak learner), the algorithm maintains two probability matrices as \( K^+_t \in {R}^{N \times L} \) and \( K^-_t \in {R}^{N \times L} \), where \( N \) is the number of data samples and \( L \) is the number of labels. These matrices store the probabilities of positive and negative residual errors (differences between student and teacher predictions).\\
The key for this algorithm is the weak learning condition that stands for finding a weak learner student for a dataset $\{(x_i, y_i)\}_{i=1}^N$ and labels $j \in \{1, 2, \dots, L\}$ that satisfies:
\begin{equation}
        \sum_{i} K^+_t(i, j) (f_t(x_i) - g(x_i))_j + K^-_t(i, j) (g(x_i) - f_t(x_i))_j > 0 \quad ,  \forall j  
    \label{eq:weaklcondition}
\end{equation}
Here,  \( f_t \) is the student model being trained, and \( g \) is the teacher model. The condition in equation~\ref{eq:weaklcondition} ensures that students improve upon the residual errors between the teacher and student models for all labels.\\

The core of the B-DISTIL algorithm is the subroutine, which searches for a weak learner. The subroutine iterates over a set of model classes \( \{F_r\} \) finding candidate model $f_t$, parametrized by $\theta$ and applying stochastic gradient descent (SGD), that minimize the following condition:
\begin{equation}
    \min_\theta \Big(- \frac{1}{\gamma}\sum_{i, j} I_{ij}^+\log \Big(1 + \frac{l(x_i)_j}{2B}\Big)
    + (1 - I_{ij}^+)\log \Big(1 - \frac{l(x_i)_j}{2B}\Big)\Big)
 \label{eq:studentloss}
\end{equation}
where $I^{+}_{ij}:= I[K^+(i, j) > K^-(i,j)]$, \( l(x_i)_j \) is the loss between the teacher and the student model at a label \( j \), and \( B \) is a constant that controls the regularization of the loss.\\

If a suitable weak learner is found, it is added to the ensemble of student models. If no weak learner is found, the algorithm expands the model class \( \{F_r\} \) and continues searching. Once a weak learner \( f_t \) is identified, the probability matrices \( K^+_t \) and \( K^-_t \) are updated to reflect the new residual errors between the teacher and student models. These updates act as the maximizer that forces the students to train in stricter learning conditions, each student ensemble gradually improves the coarse approximations of the teacher model.\\

At the end of the total number of rounds $T$, the ensemble of weak learners is composed to produce an aggregated prediction:
\[
F_T(x) = \frac{1}{T} \sum_{t=1}^{T} f_t(x)
\] 
\subsection{Mamba Architecture Theory}
The Mamba Architecture is an optimized neural network framework designed to improve the efficiency and scalability of models, particularly in sequential data tasks. It builds on traditional SSMs but introduces innovations that reduce computational overhead by selectively processing relevant information. An SSM is a mathematical framework that captures temporal dependencies by maintaining a system state $h(t)$ that evolves over time based on both previous states and current inputs $x(t)$. The general state-space representation is:
\begin{equation}
  \label{eq:SSM1}
  \begin{aligned}
    h'(t) &= \bm{A}h(t) + \bm{B}x(t) \\
    y(t) &= \bm{C}h(t) + \bm{D}x(t)
  \end{aligned}
\end{equation}
where $\bm{A}$, $\bm{B}$, $\bm{C}$, $\bm{D}$ are parameters learned by gradient descent.\\
The SSM representation in \eqref{eq:SSM1} can also be written in discrete time as: 
\begin{equation}
    \label{eq:SSM2}
    \begin{aligned}
      h_{k} &= \bm{\overline{A}} h_{k-1} + \bm{\overline{B}} x_k \\
      y_k &= \bm{\overline{C}} h_k
    \end{aligned}
\end{equation}
where $\bm{\overline{A}}$, $\bm{\overline{B}}$, $\bm{\overline{C}}$ are the discrete-time invariant state-space matrices. In this representation, the hidden state $h_k$ is recursively updated based on the previous hidden state and the current input, while the output $y_k$ is computed from the current hidden state.\\

To make this process more efficient, as shown in \citep{gu2021efficiently}, the SSM can be expressed as a convolution. By unrolling the recursion, the output $y_k$ becomes a weighted sum of the current and previous inputs, where these weights are determined by powers of the matrix $\bm{\overline{A}}$ . This leads to the construction of the convolution kernel $\bm{\overline{K}}$, which can be defined as: 
  \begin{equation}%
    \label{eq:convolution}
    \begin{aligned}
      \bm{\overline{K}} &= (\bm{C}\bm{\overline{B}}, \bm{C}\bm{\overline{A}}\bm{\overline{B}}, \dots, \bm{C}\bm{\overline{A}}^{k}\bm{\overline{B}}, \dots) 
    \end{aligned}
  \end{equation}
    \begin{equation}%
    \label{eq:convolutionxK}
    \begin{aligned}
      y &= x \ast \bm{\overline{K}}
    \end{aligned}
  \end{equation}
  where $\bm{\overline{K}}$ encodes the dynamics of the system as a series of matrix multiplications with $\bm{\overline{A}}$, $\bm{\overline{B}}$, and $\bm{\overline{C}}$. This formulation transforms the problem into a convolution of the input sequence $x$ with the kernel $\bm{\overline{K}}$ shown in \eqref{eq:convolutionxK}, thereby simplifying the computation and reducing the complexity of modeling long sequences. The matrices in \eqref{eq:convolution} are a special case of HiPPO matrices \citep{gu2020hippo} called Normal Plus Low-Rank (S4 SSMs) \citep{gu2021efficiently}, the diagonal kind of these matrices accelerate the computation of $\overline{K}$ before convolution.\\
  
For large sequences, the SSM processes all incoming data uniformly; this can lead to inefficiencies when much of the input data is irrelevant. Also, the representation depicted in equation~\ref{eq:convolution} relies on the assumption of working in a Linear Time Invariant (LTI) System, which is not always true in complex contexts. Mamba introduces Selective-State-Space Models S6 that improve traditional SSMs by selectively processing only the most relevant parts of the input sequence. The selectivity is based on the discretization method, where the continuous $(A,B)$ are transformed to their discrete version $\overline{A_k},\overline{B_k}$ through $f_A(\Delta_k, A)$ and $f_B(\Delta_k, A, B)$. Rules as the zero-order hold (ZOH) defined in the following expresions can be used as $f_A, f_B$ .
  \begin{equation}
    \label{eq:zoh}
    \bm{\overline{A_k}} = \exp(\Delta_k \bm{A})
    \qquad
  \end{equation}
\begin{equation}
    \label{eq:zoh2}
    \bm{\overline{B_k}} = (\bm{A})^{-1} (\exp(\Delta_k \bm{A}) - \bm{I}) \cdot \bm{B}
\end{equation}
Introducing time dependencies by $\Delta_k$ in the matrices seen in \eqref{eq:zoh} and \eqref{eq:zoh2} implies that the system changes to be a Linear Time Variant (LTV), which also transforms \eqref{eq:SSM2} to be selective. For instance, if $\Delta_k\rightarrow 0$ then $\overline{A_k}=I$, which makes $\overline{B_k}=0$ filtering to the input $x(k)$. On the other hand, if $\Delta_k\rightarrow\infty$ then $\overline{A_k}=0$ making the system to forget the previous state $h_{k-1}$. \\

\begin{figure}
  \begin{center}
    \includegraphics[width=\linewidth]{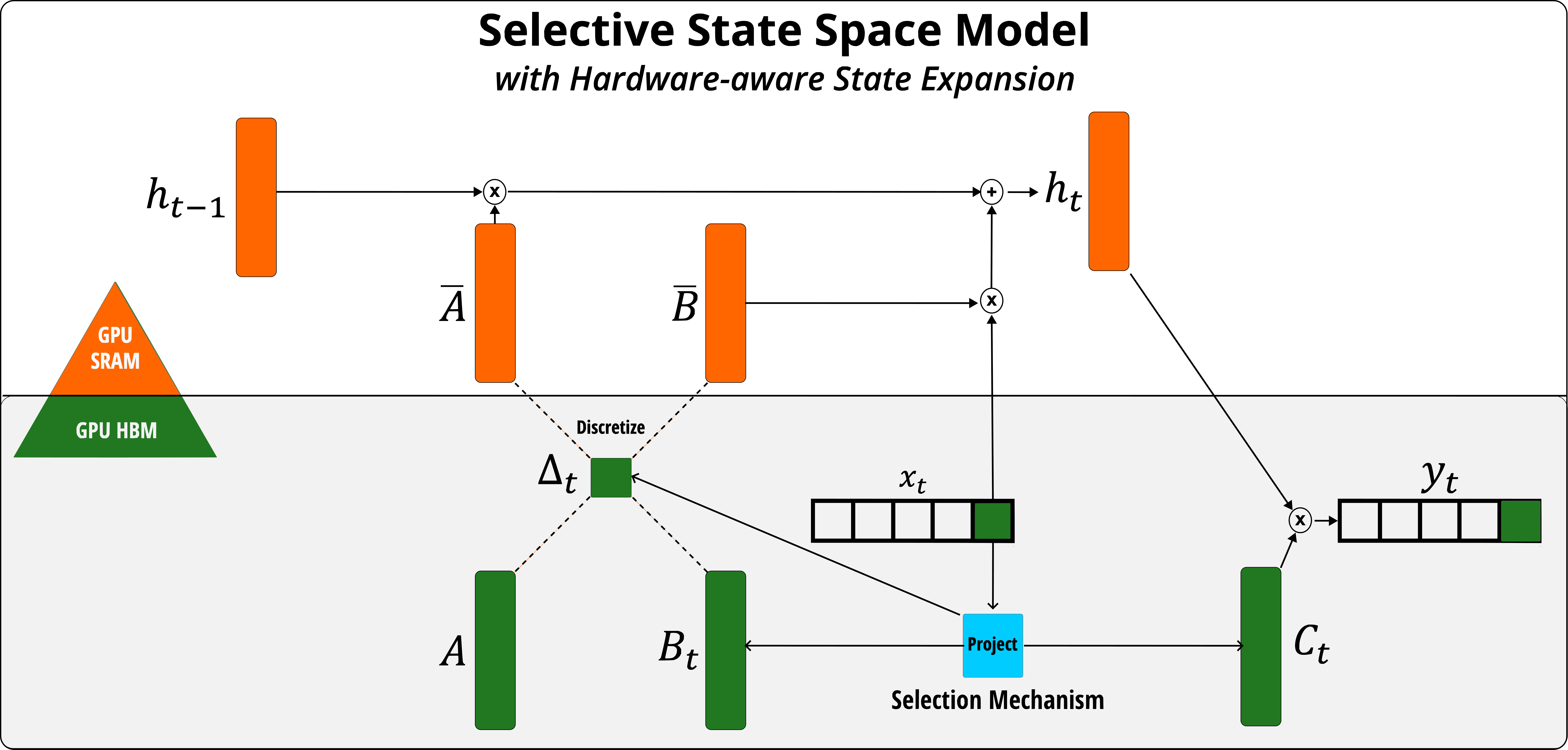}
  \end{center}
  \caption{
  Selective SSM, the flow of the data and operations, requires a hardware-aware algorithm that computes the hidden state $h$ in more efficient levels of the GPU memory hierarchy. The S-SSM is an LTVS that projects the input $x$ to construct a time dependence in $\Delta_t, B_t, C_t$. This process ensures selectivity during discretization ($A\rightarrow \overline{A}, B\rightarrow \overline{B}$) and resetting the hidden state through $C_t$ to the final output $y_t$.
}
  \label{fig:selection}
\end{figure}

The algorithm weakness of making the system LTV is that \eqref{eq:convolution} cannot be used as a simple convolution anymore; however, the method of Parallelizing Linear State Space Models, proposed by \citep{smith2022simplified}, which uses the concept of parallel scans, is implemented to efficiently compute the states of the discrete LTV SSM. The hardware-aware parallelism in Mamba further optimizes its performance by distributing computations across different hardware components, as shown in Figure \ref{fig:selection}, where the operations over the hidden state $h$ are processed at the efficient levels of the GPU memory hierarchy.\\

Mamba replaces the complex attention mechanisms and multi-layer perceptron (MLP) blocks commonly found in Transformer-based models \citep{vaswani2017attention}  with a single, unified S-SSM block. This significantly reduces the computational complexity while maintaining high performance.\\
\begin{figure*}
  \begin{center}
    \includegraphics[width=\linewidth]{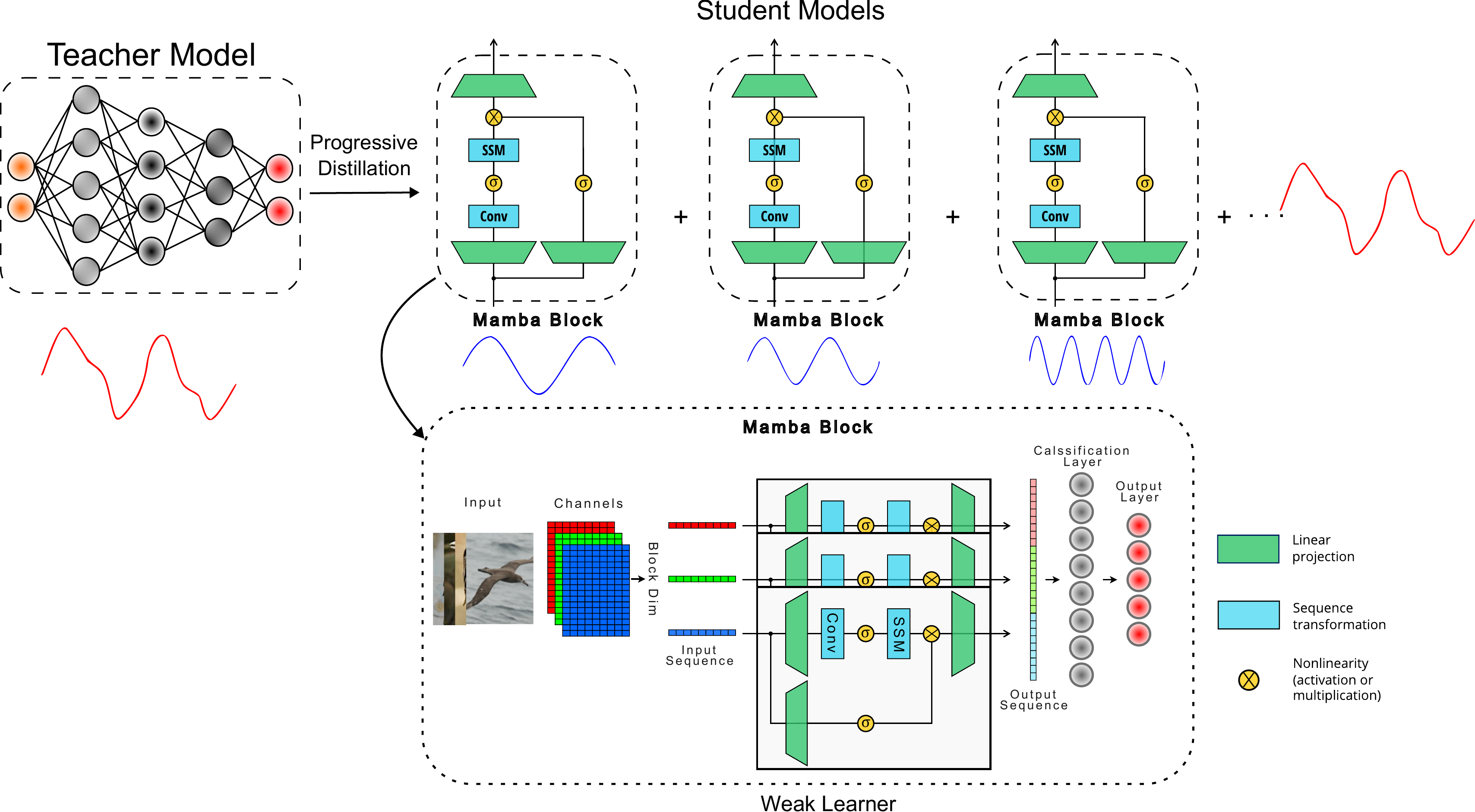}
  \end{center}
  \caption{Mamba Blocks inside Progressive Knowledge Distillation}
  \label{fig:PKDMamba}
  \end{figure*}
\subsection{Combining PKD and Mamba Architecture Methodology}
The proposed framework integrates students based on Mamba architecture within a PKD process to create an efficient model compression method that reduces computational overhead while maintaining high performance. This methodology combines the flexibility of PKD, which enables gradual model refinement, with Mamba's resource-efficient selectivity processing capabilities. Below, we break down the architecture, training pipeline, and the progressive refinement process in detail.\\

\textbf{Architecture:} At the heart of this framework, the Mamba blocks replace the standard student models used in traditional PKD (see Figure~\ref{fig:PKDMamba}). Each Mamba block serves as the fundamental building unit for the student models. These blocks incorporate S-SSM, which provide a mechanism for selective processing of input sequences, allowing the system to focus on relevant features while discarding irrelevant data.\\

The distillation process begins with a trained teacher model, which is progressively distilled into a series of student models (or weak learners) using the Mamba blocks, as illustrated in the top section of Figure~\ref{fig:PKDMamba}. Each student model is progressively refined to approximate the teacher's performance while using fewer resources. The Mamba block is key to this efficiency; it processes the input data in parallel, leveraging the inherent parallelism of the state-space matrices in the S-SSM. As described in \citep{gu2023mamba} and \citep{gu2021efficiently}, this structure allows the Mamba block to handle multiple channels of input data, extracting features from pixel sequences of an image.\\

Each input image is decomposed into several channels, and each channel is further divided into pixel sequence patches. These patches are processed as 2D sequences, each containing features and positional information. The Mamba block processes these sequences and applies selective processing to focus on the most important features. The processed sequences are passed through a series of Swish activation functions \citep{chowdhery2023palm}, which have been shown to improve model convergence and stability. At the final stage of each block, a linear layer flattens the output sequences, and a classification layer generates the final prediction logits.\\

\textbf{Progressive Knowledge Transfer:} The distillation process is progressively applied across multiple student models (see the top block sequence in Figure~\ref{fig:PKDMamba}). The first student model (the weakest learner) is trained using both true labels and the teacher's soft predictions (soft labels). This student model provides a coarse prediction, which is then refined in the subsequent students.\\

Each student model is progressively more complex than the previous one, as the weak learner condition in \eqref{eq:weaklcondition} becomes stricter, with added features in the patch sequence, a major number of Mamba blocks, and increased hidden state sizes in the S-SSM. The teacher’s knowledge is transferred not all at once, but in incremental stages, allowing each student to improve upon the last. This method ensures that the computational complexity of each student is balanced with the accuracy of the predictions. Moreover, the Mamba architecture’s ability to process multiple sequences in parallel further accelerates the distillation process, allowing the framework to operate efficiently even with complex datasets.\\

The hardware-aware Mamba algorithm plays a crucial role in this framework. It not only enables parallel processing by managing multiple student models concurrently, reducing the overall inference time, but also leverages a parallel scan operation to efficiently compute the selective processing of input sequences. This parallel scan enables simultaneous feature extraction across channels, further optimizing the overall computational flow and enhancing the system's scalability. The inference time is reduced to the time required by the most complex student plus the time to combine all student outputs. This distributed approach significantly speeds up the inference phase compared to the teacher model, making it suitable for deployment in real-time applications.\\

\textbf{Mamba Blocks and Hyperparameter Tuning:} A key advantage of Mamba architecture is its flexibility, which is achieved through careful tuning of several hyperparameters. These hyperparameters allow the student models to be tailored to specific tasks and datasets, ensuring optimal performance with minimal computational cost.\\

The following parameters were adjusted in this work:
\begin{itemize}
    \item \textbf{Number of Mamba Blocks:} Each input sequence is processed by splitting it across a series of Mamba blocks. The number of blocks is adjusted based on the complexity of the dataset. For example, more blocks are used for CIFAR-10 than for MNIST, as CIFAR-10 has more detailed image features.
    \item \textbf{Expansion Factor and features:} This factor controls how much each Mamba block expands the hidden dimensions during processing. Increasing the expansion factor allows each block to handle more complex features, improving the accuracy of the student models. Furthermore, the input can be arranged to increase the feature representation of each patch based on this expansion factor. 
    \item \textbf{Convolution Factor:} Convolutional layers within each Mamba block capture important local features from the input data. The convolution factor determines the depth and size of these layers, enabling the system to extract more detailed feature representations before passing the data to the S-SSM.
    \item \textbf{SSM Hidden State:} Each Mamba block maintains a hidden state that tracks relevant information over the input sequence. The hidden state size is adjusted based on the dataset complexity, with smaller hidden states used for MNIST and larger hidden states for CIFAR-10. This ensures that each Mamba block can handle the necessary features without overloading the system with unnecessary computations.
    \item \textbf{Hidden Layers:} After each Mamba block, the output is passed through a set of hidden layers that construct a classifier. The number of neurons in these hidden layers depends on the size of the input sequence, the number of features, and the complexity of the dataset. By carefully tuning the number of hidden layers, we balance model complexity and classification performance.
    \item \textbf{Output Neurons:} The final classification layer contains a configurable number of output neurons based on the number of classes in the dataset. For MNIST and CIFAR-10, 10 output neurons are used, as both datasets consist of 10 classes.
\end{itemize}
\textbf{Training Pipeline: } The training process follows a progressive knowledge distillation pipeline. After training the teacher model, the first student model is trained using the common KD loss seen in \eqref{eq:KDLoss}. Each subsequent student receives progressively refined knowledge from the teacher and earlier student models, gradually increasing in complexity due to a harder condition in \eqref{eq:weaklcondition}.\\

The first student model, being the weakest learner, uses a simple Mamba block for generating coarse predictions. Each subsequent student model becomes more complex by adding additional Mamba blocks, increasing the expansion factor, and enlarging the hidden state size. This progressive increase in complexity ensures that each student model can build upon the predictions of the previous students, refining the output to more closely approximate the teacher model’s performance.\\

The students are also adapted to a specific dataset; for MNIST, fewer Mamba blocks and smaller hidden state sizes are used, while for CIFAR-10, the students require more blocks and larger hidden states due to the higher complexity of the images. To validate the proposed methodology, the performance of each student model was compared to the teacher model, and we measured the accuracy, FLOPs (floating-point operations), and overall trainable parameters.

\section{Experiments}
This section describes all the hyperparameters and configurations related to the process of KD using the architecture described before. The experiments were conducted using an NVIDIA GeForce RTX 3060 GPU with 12 GB of memory. The GPU is managed using CUDA version 12.4 on a Linux-based system. 
\subsection{Datasets}
For the preliminary experiments, we used two widely recognized image classification datasets: MNIST and CIFAR-10. These datasets are commonly used as benchmarks to validate new approaches, providing a reliable foundation to test the effectiveness of PKD and Mamba before applying the framework to more complex tasks.\\

\textbf{MNIST.} This dataset \citep{lecun1998mnist} consists of 70,000 images of handwritten digits, each in grayscale and with a pixel dimension of 28 × 28. Since the images are single-channel (grayscale), the input sequence length for the Mamba architecture corresponds to 784 (28 × 28 = 784) scalar values. We used a 70-30 train-test split, resulting in 49,000 training images and 21,000 test images. The relatively small size and single-channel nature of MNIST make it a suitable dataset for validating the basic functionality of PKD and Mamba.\\

\textbf{CIFAR-10.} The dataset \citep{krizhevsky2009learning} contains 60,000 color images in 10 classes, with a pixel dimension of 32 × 32. Each image has 3 color channels (RGB), making the input sequence for Mamba correspond to 32 × 32 × 3 = 3,072 scalar values per image. The multichannel nature and higher complexity make it a good test case for evaluating the scalability of PKD in handling more sophisticated image classification tasks. The 70-30 split resulted in 42,000 training images and 18,000 test images.
\subsection{Knowledge Distillation Experiment Setup}
For PKD, we trained a sequence of seven student models, each learning incrementally from the teacher model. The key steps in the training process were as follows:\\

\textbf{Training Parameters: } We used the following training parameters for the experiments:
\begin{itemize}
    \item{\textbf{Learning rate:}} The learning rate was set to 0.0001, and a learning rate decay was applied to reduce it gradually during training.
    \item{\textbf{Batch size:}} A batch size of 32 was used to balance memory efficiency and training speed.
    \item{\textbf{Epochs:}} Each student model was trained for 50 epochs, with early stopping criteria applied if the validation loss did not improve for 5 consecutive epochs. Also, after every 10 epochs, a weak learner validation was applied to stop the training.
    \item{\textbf{Loss function:}} For distillation, the loss function used a combination of cross-entropy distillation loss for the soft labels generated by the teacher model. We used a temperature of 2. The PKD loss function was configured in the second term, as was seen in \eqref{eq:studentloss}, to give a student the adaptability to enforce the coarse prediction.  
\end{itemize}

\textbf{Teacher training: } The teacher model was first trained on the full training dataset using the standard cross-entropy loss function and backpropagation. The architecture of both teachers (MNIST and CIFAR-10) is the same as that explained in Figure \ref{fig:PKDMamba} for the students. However, complex hyperparameters for the teacher were chosen to generate good accuracy. For MNIST, the teacher reaches an accuracy of 98\% and 87\% for CIFAR-10.\\

\textbf{Evaluation: }To evaluate our model we relied on three major metrics, accuracy to get an idea about the performance, model Size reflected by the number of parameters, and floating point operations per seconds (FLOPs) for measuring computational efficiency.

\section{Results and Discussion}
In this section, we present the preliminary results from the experiments conducted using  MNIST and CFAR-10 datasets, which evaluated the performance of both teacher model and student models generated through the PKD process. The experiment aimed to demonstrate the efficiency of PKD combined with mamba blocks by analyzing computational cost in terms of FLOPs and accuracy.
\subsection{Computational Efficiency}
Table \ref{tab:mnistPKD} shows the general architecture of the MNIST Classifier, with the transformation dimensions and parameters for the teacher and student models. The teacher model consists of 28 Mamba blocks with an SSM dimension of 128, resulting in a total of 34,674 parameters and 94,684 FLOPs. In contrast, the student models were progressively increasing in size by adjusting the number of blocks and the state dimensions. The smallest student model (Student~1) uses only 1 block and has a state dimension of 16, resulting in just 620 parameters and 686 FLOPs, a significant reduction in computational cost compared to the teacher model.\\

\begin{figure}
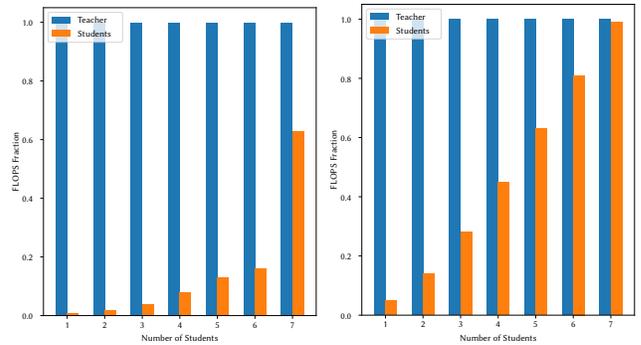

    \centering
    \begin{subfigure}{0.49\linewidth}
        \centering
        \resizebox{\textwidth}{!}{\input{images/BAR_MNIST.pgf}}
        \caption{MNIST}
    \end{subfigure}
    \begin{subfigure}{0.49\linewidth}
        \centering
        \resizebox{\textwidth}{!}{\input{images/BAR_CFAR_10.pgf}}
        \caption{CIFAR-10}
    \end{subfigure}
    \caption{Performance Teacher Students Comparison}
    \label{fig:FlopsStudents}
\end{figure}

The bar chart presented in Figure~\ref{fig:FlopsStudents} further highlights the performance improvements achieved by the student models. As the number of student models increases, the FLOPs fraction (relative to the teacher model) increases significantly, mainly because each further student has a stricter weak learner condition (see equation~\ref{eq:weaklcondition}). The largest student model (Student~7) retains 55\% of the teacher’s FLOPs, whereas the smallest model (Student~1) only requires about 1\% of the teacher's computational resources. Student~1 has a clear reduction in the computational cost while maintaining acceptable accuracy, which demonstrates the efficiency gained through the distillation process.\\

\begin{table}
\caption{MNIST overview of model configurations and some performance metrics, including the number of blocks, state dimension, total parameters, FLOPs, and accuracy for each model variant.} 
\label{tab:mnistPKD} 
\begin{tabular}{c  r  r  r  r r} 
\toprule 
\textbf{Model} & \textbf{N.} & \textbf{State} & \textbf{Total} & \textbf{Flops} & \textbf{Acc} \\
 &\textbf{Blocks}&\textbf{Dim.} & \textbf{Parameters}& & \\
\hline
\textbf{Teacher}& 28  & 128 & \textbf{34,674} & \textbf{94,684} & \textbf{98\%} \\
\hline
\textbf{Student 7}& 14 & 32 & 11,966 & 51,744& \textbf{98\%} \\ 
\textbf{Student 6}& 7  & 64 & 5,078 & 7,448 & \textbf{93\%} \\  %83-4
\textbf{Student 5}& 7  & 32 & 3,734 & 7,448 & \textbf{91\%} \\    %85-4
\textbf{Student 4}& 4  & 64 & 3,674 & 5,096 & \textbf{88\%} \\    %85-4
\textbf{Student 3}& 1  & 64 & 2,378 & 2,744&  \textbf{84\%} \\    %82-4
\textbf{Student 2}& 1  & 32 & 1,206 & 1,372& \textbf{76\%} \\    %79-8
\textbf{Student 1}& 1  & 16 & 620 & 686 & \textbf{60\%} \\       %72-16

\multicolumn{3}{c}{\textbf{All 7 Students}} & \textbf{28,656} &\textbf{76,538}& \textbf{98\%} \\
\bottomrule
\end{tabular} 
\end{table}
For CIFAR-10 Table \ref{tab:CIFARPKD} shows the architecture and performance parameters for the model, which was trained using 96 mamba blocks with a state dimension of 256, resulting in a total of 443,530 parameters and 2,390,016 FLOPs. In contrast, the student models progressively increase in complexity by adjusting the number of blocks and state dimensions. The weak learner 1 uses 24 blocks and has a state dimension of 128, resulting in 22,177 parameters and 119,506 FLOPs, which marks a significant reduction in computational cost compared to the teacher model, consuming around 5\% of teacher FLOPs. However, for this dataset, the first student coarse prediction is only 50\% accurate. On the other hand, the largest student model retains approximately 19\% of the teacher’s FLOPs, using 96 blocks and a state dimension of 128. \\

\begin{table}
\caption{CIFAR-10  overview of model configurations and some performance metrics, including the number of blocks, state dimension, total parameters, FLOPs, and accuracy for each model variant.} 
\label{tab:CIFARPKD} 
\begin{tabular}{ c  c  r  r  r r} 
\toprule 
\textbf{Model} & \textbf{N.} & \textbf{State} & \textbf{Total} & \textbf{Flops} & \textbf{Acc} \\
&\textbf{Blocks}&\textbf{Dim.} & \textbf{Parameters}& & \\
\hline
\textbf{Teacher}& 96  & 256 & \textbf{443,530} & \textbf{2,390,016} & \textbf{87\%} \\
\hline
\textbf{Student 7}& 96 & 128 & 83,528 & 450,106& \textbf{86\%}\\ %93
\textbf{Student 6}& 96  & 128 & 78,730 &  430,340& \textbf{84\%}\\  %83-4
\textbf{Student 5}&48  & 256 & 76,908 & 420,202 &\textbf{81\%} \\    %85-4
\textbf{Student 4}&48  & 256 & 72,214 & 405,304 & \textbf{78\%}\\    %85-4
\textbf{Student 3}&48  & 128 & 65,242 & 351,560 & \textbf{73\%}\\    %82-4
\textbf{Student 2}&24  & 256 & 38,224&  206,080& \textbf{65\%}\\    %79-8
\textbf{Student 1}&24  & 128 & 22,176 & 119,506 & \textbf{50\%}\\       %72-16
\multicolumn{3}{c}{\textbf{All 7 Students}} & \textbf{437,022} &\textbf{2,383,098} & \textbf{86\%} \\
\bottomrule
\end{tabular} 
\end{table}
The progressive nature of the distillation process allows each student model to capture a different level of complexity from the teacher model. As seen in the Table \ref{tab:CIFARPKD}, the students’ number of parameters and FLOPs gradually increase from, 22,176 parameters to 83,528 parameters. We can also say that the first 3 students have the largest improvement in performance; the coarse prediction goes from 50\% to 73\%.  From here, the following students only refine a bit of the prediction but consume more resources than the first models. This setup should be controlled because we also expect to pay a computational cost for each further student that does not exceed the computational resource of the teacher.\\

The bar chart (Figure \ref{fig:FlopsStudents}) further illustrates the FLOPs fraction relative to the teacher model. The small student model uses only 5\% of the teacher's computational resources, while the largest student model almost reaches 100\% of the teacher’s computational load. Despite the reduced computational complexity of the first students with a coarse prediction of 73\%, the model as a group does not reach some effectiveness of the progressive knowledge distillation process. The significant increase in computational resources for the last stages of distillation confirms that the algorithm must be relaxed in the weak learner's condition.
\begin{figure}
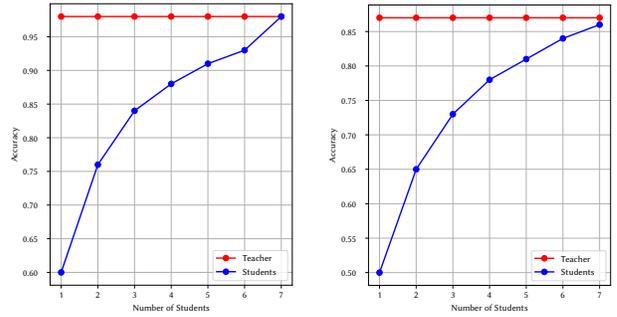

    \centering
    \begin{subfigure}{0.49\linewidth}
        \centering
        \resizebox{\textwidth}{!}{\input{images/Figure_1.pgf}}
        \caption{MNIST}
    \end{subfigure}
    \begin{subfigure}{0.49\linewidth}
        \centering
        \resizebox{\textwidth}{!}{\input{images/Figure_2.pgf}}
        \caption{CIFAR-10}
    \end{subfigure}
    \caption{Accuracy vs. Number of Students}
    \label{fig:AccStudents}
\end{figure}

\subsection{Model Performance}
In terms of accuracy, the teacher model achieved the highest performance on the MNIST test set, as expected. However, the student models, even with significantly fewer parameters, managed to retain a considerable portion of the teacher’s accuracy. As shown in Table \ref{tab:mnistPKD} and Figure \ref{fig:AccStudents}, which presents the accuracy of the teacher and student models, the largest student (Student 7) achieved a close approximation of the teacher's accuracy, with the accuracy decreasing slightly for smaller student models.
The progressive nature of the distillation process enabled smaller models like Student 1 to achieve 63\% accuracy, while Student 7 achieved approximately 98\% accuracy, almost equal to the teacher's performance. This gradual reduction in both FLOPs and accuracy highlights the flexible trade-off between computational cost and model performance, making it possible to choose a student model based on the available computational resources.\\

In the case of the CIFAR-10 dataset, the performance trend of the student models differs from that observed with the MNIST dataset. The teacher model achieved an accuracy of 87\%, serving as the benchmark for student models. As detailed in Table, \ref{tab:CIFARPKD} and illustrated in Figure \ref{fig:AccStudents}, resembling the largest student model (Student~7) attained an accuracy of 86\%, which is a 1\% decrease compared to the teacher. It can be seen as a good result in KD process; however, this slight reduction in error comes at the cost of utilizing approximately 99\% of the teacher's computational resources, which is significantly higher than the resource usage of all student models in the MNIST experiments (63\% see Figure \ref{fig:FlopsStudents}).\\

Unlike MNIST, where smaller student models could match the teacher's accuracy with minimal computational resources, the students in CIFAR-10 models exhibited a more substantial drop in accuracy for small students' sizes. For instance, Student~1, the smallest model, achieved only 50\% accuracy with 5\% of computational resources, which is a considerable decline from the teacher's performance. The notable gap in performance between the teacher and smaller student models on CIFAR-10 can be attributed to the dataset's inherent complexity. CIFAR-10 comprises colored images with diverse classes and intricate features, requiring models with a higher capacity to capture the nuanced patterns effectively. Smaller student models with limited parameters struggle to learn these complex representations, leading to a significant drop in accuracy.\\

Finally, even though the performance of students as a group seems compromised compared to the teacher's performance, the PKD approach remains advantageous for CIFAR-10. Each student model can operate in parallel within the final ensemble, meaning that the overall response time is determined by the largest student model. With the largest student utilizing approximately 19\% of the teacher's FLOPs, this sets a reasonable maximum response time for the entire system. Consequently, Progressive Knowledge Distillation proves to be an effective strategy for CIFAR-10, with multiple student models progressively approaching the teacher's performance while significantly reducing computational costs.
\section{Conclusion and Future Work}
As the demand for real-time, resource-efficient machine learning models grows, the combination of Progressive Knowledge Distillation and Mamba Architecture offers a promising solution for achieving high-performance models that can be deployed on resource-constrained devices. Research in adaptive neural networks, early-exit mechanisms, and selective attention mechanisms all complement the selective processing and progressive learning principles inherent in Mamba and PKD.\\

This work is a preliminary result of how Mamba can be adapted to other techniques of Knowledge Distillation. As was explained before, PKD has the advantage of working at the first stages with a good approximation of the teacher without paying too much computational cost. However, knowledge distillation reaches a limit as we go further, finding weak learners and imposing a harder constraint on students. In the case of MNIST, the limit for the architecture that we proposed has not yet been reached. Meanwhile, CIFAR-10 reaches that limit using the same 7 students. \\

The flexibility of Mamba architecture in enabling the configuration of their hyperparameters and the introduction of more blocks improves how we construct students as weak learners. In \cite{dennis2023progressive}, the author explained that one constraint is that the algorithm must consider pre-define a set of student classes \{$\mathcal{F}$\}, and the way to construct these classes must be connected to the teacher architecture, expanding the architecture from a small base model. Mamba architecture, without any additional set of classes configured, allows us to propose a candidate student easily by only changing the number of blocks or slightly increasing the dimension hidden state for an SSM. The number of independent blocks in Mamba also allows for increasing complexity, going from a single-channel dataset (MNIST) to a multichannel dataset (CIFAR-10), managing the new channel sequences as new blocks of Mamba. \\

Additionally, future work could explore combining these techniques with Neural Architectures that are optimized to extract visual features, especially for image classification. Increasing the number of sets of classes \{$\mathcal{F}$\} for students could relax the weak learner condition for larger and more complex datasets than those analyzed in this paper. The loss function is another characteristic that must be adapted to these newly proposed architectures. In the case of CIFAR-10 in this work, the learning condition forces the students to be more complex in their architecture, only paying attention to how this student can improve the coarse prediction. However, the concept of KD goes beyond the perception of accuracy; the KD process must consider how well the student as an individual is able to fit the knowledge from the teacher and then how well it fits its knowledge in the progressive group.   

\section*{Acknowledgments}
This work was supported by the collaboration project LLTAT21278 with Bolt Technologies.

\bibliographystyle{ACM-Reference-Format}

\end{document}